\documentclass{ieice-tecrep}
\usepackage{graphicx}
\usepackage[fleqn]{amsmath}
\usepackage{newtxtext}
\usepackage[varg]{newtxmath}
\usepackage{algorithm2e}
\usepackage{natbib}
\usepackage{hyperref}
\usepackage{booktabs}

\tecrepyear{2026}
\title{Decentralized Clustered Federated Learning for Cellular Traffic Forecasting over LEO Constellations}
\subtitle{}
\authorlist{%
\authorentry{Jonas Kirch}{tohoku}
\authorentry{Tiago Koketsu Rodrigues}{tohoku}
\authorentry{Yuichi Kawamoto}{tohoku}
}
\affiliate[tohoku]{Tohoku University, Graduate School of Information Science \ 2-1-1 Katahira, Aoba-ku, Sendai, Miyagi 980-8577}

\breakauthorline{3}

\begin{document}

\begin{abstract} 
Low Earth Orbit (LEO) satellite constellations are positioned to act as orbital edge-compute nodes in emerging 6G non-terrestrial networks, supporting forecasting and control tasks for ground cellular infrastructure. However, training such models is challenged by heterogeneous ground-cell traffic patterns, intermittent links to ground stations, and the satellites' own orbital motion, which continuously shifts the data each satellite observes. In this work, we extend Decentralized Federated Clustering (DFCA) \cite{Kirch2025DFCA} to the LEO setting and apply it to cellular internet-traffic forecasting on the Telecom Italia Milan+Trentino dataset. Satellites cooperate exclusively through $k$-nearest inter-satellite gossip with no ground-side aggregator, while zero-shot head selection lets each satellite specialize to the latent traffic regime of the ground cell currently beneath it. Averaged over $K{=}20$ post-training orbital snapshots, our extension reduces forecasting Root Mean Squared Error (RMSE) by $15.3\%$ over a decentralized FedAvg baseline of equivalent capacity, with cluster recovery accuracy of $85.3\%$. The results indicate that clustered specialization yields measurable gains under the structural heterogeneity inherent to realistic cellular-over-LEO scenarios. 
\end{abstract}

\maketitle

\section{Introduction}
\label{sec:intro}

Cellular traffic forecasting is a long-standing problem in mobile network management, used to drive resource allocation, slicing, and capacity planning. With the standardization of non-terrestrial networks (NTN) in 3GPP Release 17 and the trajectory toward 6G satellite mobile edge computing (MEC) integration~\cite{Lin2019NTN} \cite{Pachler2021}, an emerging research direction places forecasting tasks directly on Low Earth Orbit (LEO) satellites that observe ground cells as part of their orbital footprint. This shifts the problem from a centralized terrestrial setting to one with three distinguishing properties.

\begin{enumerate}
    \item \textbf{No persistent ground link}: Satellites cannot rely on a ground-based parameter server for synchronous aggregation. Ground-station visibility is intermittent and bandwidth-limited \cite{Razmi2022GroundAssisted} \cite{Yang2024DFedSat}. Decentralized federated learning over inter-satellite links (ISLs) is therefore the natural fit.
    \item \textbf{Orbital motion induces continuous data shift}: 
    The ground cell under a given satellite changes as the satellite orbits, so each satellite's effective training distribution is time-varying. This distinguishes the LEO setting from static decentralized federated learning, where each client's data is fixed.
    \item \textbf{Structural inter-regional heterogeneity}: Cellular traffic patterns differ qualitatively across regions, with dense urban cells exhibiting smooth, sustained internet-data peaks, while rural cells display sparse, bursty patterns~\cite{Pavlidis2024FL}. These are not merely magnitude differences but differences in temporal shape that a single shared model cannot resolve without internal capacity waste.
\end{enumerate}

In a prior work \cite{Kirch2025DFCA} we introduced \emph{Decentralized Federated Clustering} (DFCA), a fully decentralized clustered federated learning algorithm in which clients maintain $p$ candidate heads and select among them via zero-shot loss comparison, exchanging only same-head weights through neighbor gossip. DFCA was originally evaluated on standard FL benchmarks with static client graphs.

This paper extends DFCA to the LEO satellite scenario described above. Specifically:

\begin{enumerate}\setlength\itemsep{1pt}
\item We adapt DFCA to a $k$-nearest-neighbors inter-satellite gossip topology that is recomputed periodically from orbital geometry.
\item We expose each satellite to a time-varying local data distribution.
\item We evaluate on cellular internet-traffic forecasting using the Telecom Italia Milan+Trentino dataset in a setting where the target dynamics differ structurally between regions.
\end{enumerate}

\begin{figure}
    \centering
    \includegraphics[width=\linewidth]{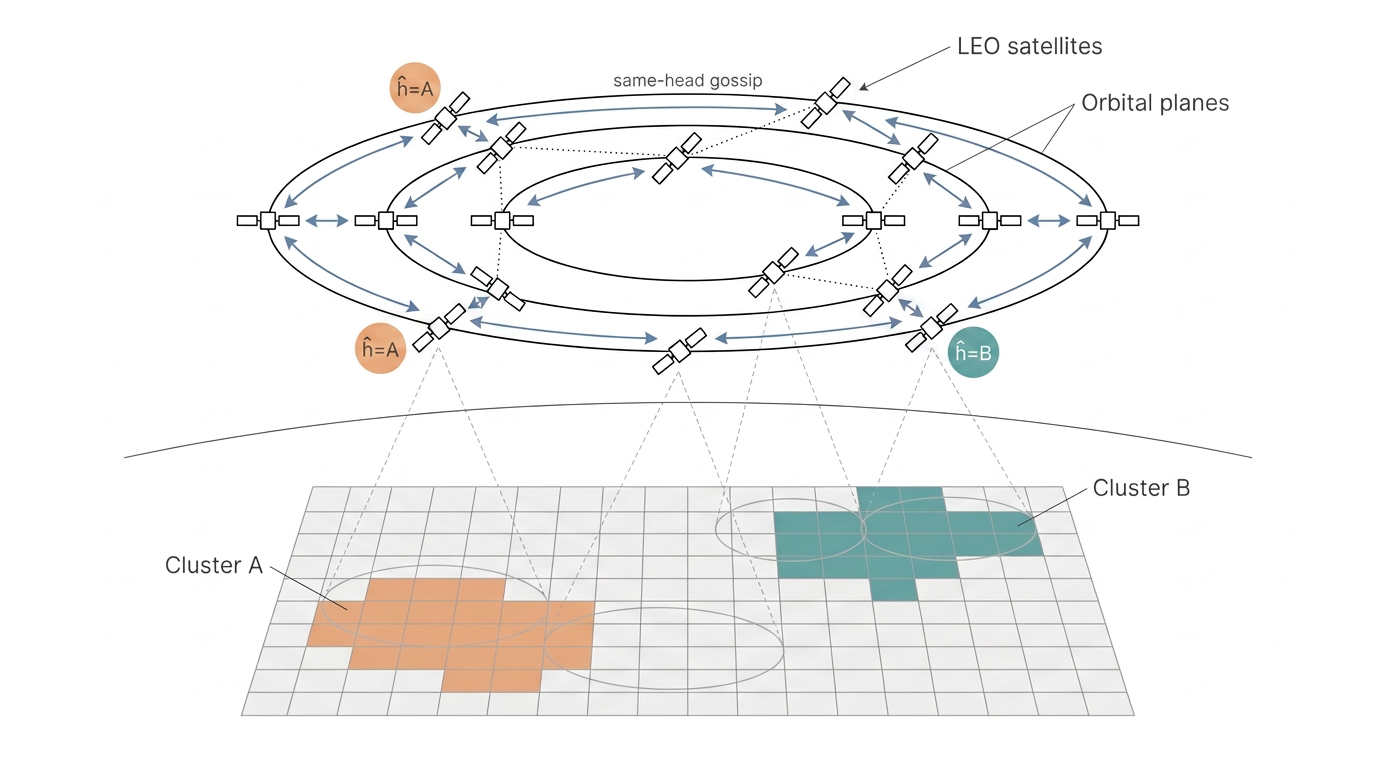}
    \caption{Experimental Scenario for Milan/Trentino Dataset with LEO-Satellites}
    \label{fig:scenario}
\end{figure}

The remainder of this paper is organized as follows: Section~\ref{sec:related} reviews related work in clustered and satellite federated learning. Section~\ref{sec:method} describes the methodology and presents the adapted algorithm. Section~\ref{sec:experiments} reports experimental results, and Section~\ref{sec:conclusion} contains a short discussion and a conclusion of our work.

\section{Related Work}
\label{sec:related}

\subsection{Clustered Federated Learning}
Clustered Federated Learning (CFL) has emerged as an effective paradigm to address statistical heterogeneity (non-IID data) by grouping clients with similar data distributions. The Iterative Federated Clustering Algorithm (IFCA) \cite{Ghosh2020IFCA} introduced the foundational cluster-identity estimation and head-selection mechanisms that this work inherits. While \cite{Marfoq2021FedEM} generalized hard cluster assignment to soft multi-task mixture models, these approaches assume a centralized parameter server to coordinate model routing and cluster assignments.

To eliminate the central server bottleneck, recent works have proposed fully decentralized, peer-to-peer, clustered and personalized federated architectures. For instance, Lin \textit{et al.} \cite{Lin2025FedSPD} proposed FedSPD, a soft-clustering decentralized federated learning algorithm designed to achieve consensus on distinct cluster models over low-connectivity graphs. Similarly, P3P-Fed \cite{P3PFed2025} introduces node-level decentralized clustering with personalized head training to bypass central aggregation bottlenecks altogether. Under dynamic environments, static cluster configurations degrade due to real-time distribution shifts. To combat this, adaptive hierarchical clustering strategies \cite{RMPFD2025} and dynamic round-wise client clustering methods \cite{DCPFL2025} have been introduced. However, both rely on a central server that observes model updates or training-loss statistics from a largely stable client population before deciding how to (re-)cluster: RMPFD's \cite{RMPFD2025} clustering manager operates on client models collected at a cloud aggregator, while DC-PFL's \cite{DCPFL2025} re-clustering trigger is computed from the average training loss across all clients each round and depends on model discrepancy stabilizing over a sustained window per client. Neither assumption holds in the LEO setting, where no ground-side aggregator is available to gather synchronous, population-wide statistics, and where a satellite's underlying data distribution itself drifts continuously with its orbital footprint rather than remaining fixed long enough for server-side discrepancy tracking to converge.

The Decentralized Federated Clustering Algorithm (DFCA) \cite{Kirch2025DFCA} removes the central coordinator via sequential running-average aggregation over local neighbor gossip. However, DFCA is formulated for relatively static network topologies. Our work directly extends this line of research, adapting the decentralized clustering mechanism to the rapidly shifting, time-varying network graphs inherent to dynamic LEO satellite constellations.

\subsection{Federated Learning on LEO Satellites}
Early implementations of federated learning (FL) in LEO satellite constellations relied heavily on centralized architectures, such as the ground-assisted aggregation framework introduced by Razmi \textit{et al.} \cite{Razmi2022GroundAssisted}. Rodrigues \textit{et al. }\cite{Rodrigues2023} similarly analyze the trade-off between centralized and distributed learning costs in satellite networks, proposing a hybrid scheme that adaptively switches between the two on a cloud-equipped satellite. However, due to intermittent ground station connectivity and high uplink latency, recent literature has shifted toward utilizing ISLs to enable decentralized federated learning (DFL). 

Within this paradigm, Zhai \textit{et al.} \cite{Zhai2024FedLEO} proposed an offloading-assisted DFL framework, while Yang \textit{et al.} \cite{Yang2024DFedSat} and Zhou \textit{et al.} \cite{Zhou2024} focused on improving communication efficiency and robustness against dynamic, time-varying inter-orbit ISL topology changes. To secure these decentralized updates and mitigate Byzantine threats in orbital environments, Zhao \textit{et al.} \cite{Zhao2026} designed a secure coordination scheduling scheme, while Xu \textit{et al.} \cite{XU2025} drastically compressed communication overhead by sharing distilled datasets rather than raw weights over the ISL network. 

While these DFL schemes successfully solve the physical routing and link-dynamics challenges of LEO constellations, they typically optimize for a single, globally shared model. Consequently, they do not address the spatial-temporal clustered heterogeneity of the surface environments over which they transit.

\subsection{Federated Traffic Forecasting} Predicting spatial-temporal cellular traffic is critical for proactive resource management, but doing so via centralized machine learning introduces massive telemetry overhead. On terrestrial infrastructure, Pavlidis \textit{et al.} \cite{Pavlidis2024FL} provided a comprehensive baseline evaluation of FL for cellular traffic forecasting using real-world public data, such as the multi-source Telecom Italia Milan/Trentino dataset \cite{Barlacchini2015}. However, their evaluation assumes a fixed client-to-cell mapping: Each terrestrial client trains on a consistent, unchanging subset of the traffic data throughout the FL process. This assumption breaks down in the orbital setting, where a satellite's client identity is decoupled from any single ground cell. As the satellite transits, it is successively exposed to the traffic patterns of whichever cell currently lies within its footprint, so the same physical client must generalize across regimes it was never continuously exposed to during training rather than specialize to one. Their protocol therefore offers a useful forecasting benchmark and dataset but not a framework for this time-varying client-data relationship, which motivates the orbital snapshot evaluation and clustering mechanism we introduce in Section~\ref{sec:method}.

Conducting traffic forecasting directly on LEO satellites introduces unique challenges due to dynamic beam coverage and changing user densities. To address this, Jiang \textit{et al.} \cite{Jiang2024} proposed an edge-based FL architecture for traffic prediction in satellite-terrestrial integrated networks (STINs). Furthermore, Zeydan \textit{et al.} \cite{zeydan2025FedKan} explored advanced architectures like Kolmogorov-Arnold Networks (Fed-KAN) to reduce complexity in dynamic non-terrestrial environments, while Liang \textit{et al.} \cite{LIANG2026} designed resource-efficient, lightweight transformers to adapt to the tight hardware budgets of LEO edge nodes. 

None of these current approaches, however, account for the fact that a transiting satellite's observed data distribution shifts continuously as it flies over distinct, heterogeneous ground-cell environments. This paper bridges this gap by extending decentralized federated clustering (DFCA) \cite{Kirch2025DFCA} to allow satellites to cooperatively self-organize into specialized sub-clusters on the fly.

\section{Methodology}
\label{sec:method}
This section explains the methodology. While this section explains parts of our algorithm in more detail, the full algorithm can be seen in \ref{alg:dfca-leo}. Additionally, scenario is illustrated in Figure \ref{fig:scenario}.

\subsection{System Model}
\label{sec:sys}

We consider a constellation of $m$ active LEO satellites distributed across $P$ orbital planes, observing a ground region partitioned into $G$ cells. At each training epoch $t$, satellite $i$ is assigned a single ground cell $c_i^{(t)} \in \{1,\dots,G\}$ from which it draws $n$ training samples $(X, y)$, where $X$ represents data and $y$ represents labels. The cell sequence $(c_i^{(0)}, c_i^{(1)}, \dots)$ is determined by the satellite's orbital trajectory.  The set of ground cells is partitioned (latently) into $p$ \emph{regimes} corresponding to qualitatively different traffic patterns. Regime identity is not observed by the satellites and must be recovered from data.

\subsection{Communication Topology}

Satellites communicate exclusively via ISLs, while no ground station participates in training. At regular intervals $\Delta_{\text{re}}$, the communication graph is rebuilt from the current orbital positions: for each satellite $i$, we identify the $k$ satellites closest in Euclidean distance and add bidirectional edges. The resulting graph is sparse and time-varying.

\subsection{Local Forecasting Model}

Each satellite $i$ maintains $p$ candidate cluster heads $\{\theta_{i,1}, \dots, \theta_{i,p}\}$, where each $\theta_{i,k}$ parameterizes a two-layer Long Short-Term Memory (LSTM) with hidden dimension~64, mapping an input sequence $X \in \mathbb{R}^{\ell \times d}$ to a forecast $\hat{y} \in \mathbb{R}^{h}$ with lookback $\ell$ and horizon $h$.

\subsection{Zero-Shot Head Selection}

At each epoch, satellite $i$ selects the head that best matches its current local data via a variance-normalized mean-squared error:
\begin{equation}
\hat{h}_i = \arg\min_{k \in \{1,\dots,p\}}
\frac{\mathrm{MSE}(f_{\theta_{i,k}}(X),\,y)}
     {\mathrm{Var}(y) + \varepsilon}.
\label{eq:headpick}
\end{equation}
Within a single selection step, the denominator is constant across candidate heads, so the choice of $\hat{h}_i$ is identical to that of raw MSE selection. The variance normalization renders the score \emph{comparable across batches and satellites}, which we exploit for diagnostic logging and for forward compatibility with thresholded extensions.

\subsection{Local Update and Sequential Gossip}

Local optimization follows standard mini-batch stochastic gradient descent (SGD), in which
each parameter update moves opposite the gradient of the loss evaluated on a sampled batch,
scaled by a learning rate $\eta_t$:
\begin{equation}
    \theta \leftarrow \theta - \eta_t \nabla_\theta \mathcal{L}(\theta; \mathcal{B}),
\end{equation}
where $\mathcal{B} \subset (X, y)$ denotes a mini-batch drawn from satellite $i$'s local
dataset. After head selection, satellite $i$ applies this update for $\tau$ steps to
$\theta_{i,\hat{h}_i}$:
\begin{equation}
    \theta_{i,\hat{h}_i} \leftarrow
        \theta_{i,\hat{h}_i} - \eta_t \nabla_\theta
        \mathrm{MSE}(f_{\theta_{i,\hat{h}_i}}(X), y).
\end{equation}
The other $p-1$ heads remain frozen for this epoch. Updated weights are then exchanged with
neighbors in the current ISL graph. For each incoming neighbor weight $\theta_{j,\hat{h}_j}$
with $\hat{h}_j = \hat{h}_i$, satellite $i$ applies a sequential running average:

\begin{equation}
\theta_{i,\hat{h}_i}
\;\leftarrow\;
\frac{r}{r+1}\,\theta_{i,\hat{h}_i}
\;+\;
\frac{1}{r+1}\,\theta_{j,\hat{h}_j},
\label{eq:seqagg}
\end{equation}
where $r$ is incremented each time a same-head neighbor weight has been merged this epoch. This avoids buffering and gives a communication-efficient alternative to batch averaging~\cite{Kirch2025DFCA}.

\begin{algorithm}[t]
\DontPrintSemicolon
\SetAlgoLined
\KwIn{satellites $i = 1, \dots, m$;
      heads per satellite $p$;
      local steps $\tau$;
      gossip neighbors $k$;
      total epochs $T$;
      re-graph interval $\Delta_{\text{re}}$.}
\KwOut{personalized weights $\theta_{i,\hat{h}_i}$ per satellite.}
\BlankLine
Initialize $\{\theta_{i,k}\}$ for all $i$, $k$\;
\For{$t = 1$ \KwTo $T$}{
  \If{$t \bmod \Delta_{\text{re}} = 0$}{
    Rebuild $k$-NN ISL graph $\mathcal{N}_i^{(t)}$ from orbital positions\;
  }
  \ForEach{satellite $i$ in parallel}{
    Observe current cell $c_i^{(t)}$ and sample
    $(X, y) \sim \mathcal{D}_{c_i^{(t)}}$\;
    \tcp{Zero-shot head selection, Eq.~\eqref{eq:headpick}}
    $\hat{h}_i \leftarrow
       \arg\min_k
       \mathrm{MSE}\!\bigl(f_{\theta_{i,k}}(X),\,y\bigr)
       / (\mathrm{Var}(y) + \varepsilon)$\;
    \tcp{Local SGD on selected head}
    \For{$s = 1$ \KwTo $\tau$}{
      $\theta_{i,\hat{h}_i} \leftarrow
        \theta_{i,\hat{h}_i} - \eta_t \nabla_\theta
        \mathrm{MSE}(f_{\theta_{i,\hat{h}_i}}(X), y)$\;
    }
    $r \leftarrow 1$\;
    \tcp{Sequential gossip aggregation, Eq.~\eqref{eq:seqagg}}
    \ForEach{$j \in \mathcal{N}_i^{(t)}$
             \textbf{with} $\hat{h}_j = \hat{h}_i$}{
      $\theta_{i,\hat{h}_i} \leftarrow
        \frac{r}{r+1} \theta_{i,\hat{h}_i}
        + \frac{1}{r+1} \theta_{j,\hat{h}_j}$\;
      $r \leftarrow r + 1$\;
    }
  }
}
\caption{Algorithm (per-satellite view)}
\label{alg:dfca-leo}
\end{algorithm}

\section{Experiments}
\label{sec:experiments}
As proof of concept, we conducted experiments on the MNIST dataset, introducing heterogeneity via rotation of the data, as in Ghosh \textit{et al.} \cite{Ghosh2020IFCA} and Kirch \textit{et al.} \cite{Kirch2025DFCA}. Fitting to the scenario, we also experimented with the commonly used Milano+Trentino Telecommunications dataset \cite{Barlacchini2015}, applying the system model and topology described in Section~\ref{sec:method}. Here, the heterogeneity is mainly formed by regional differences in the data, as demonstrated in Figure~\ref{fig:data}. Additionally, we introduced slightly more heterogeneity by splitting the dataset into voice and internet communication.

\begin{figure}[!h]
    \centering
    \includegraphics[width=\linewidth]{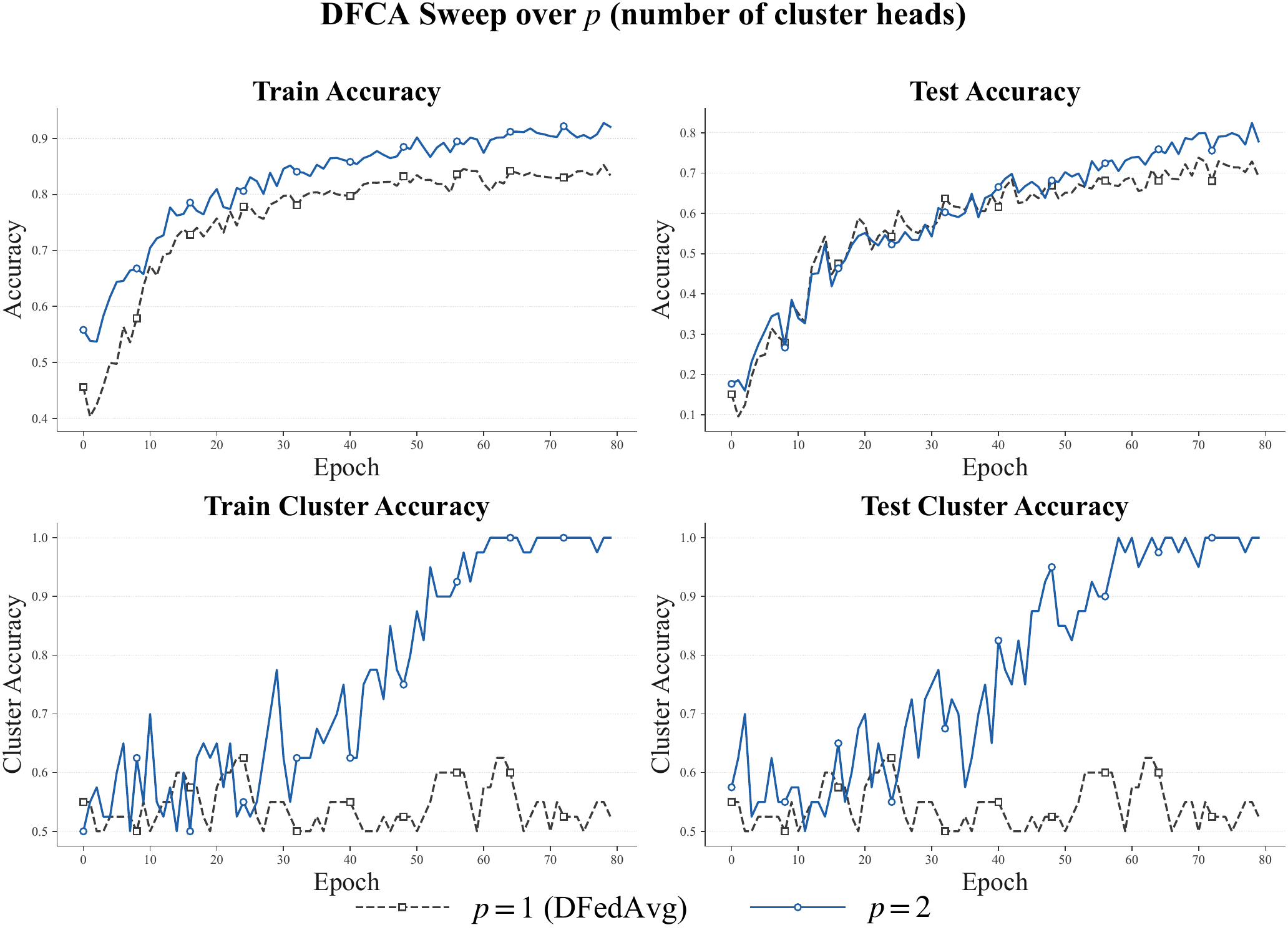}
    \caption{Training statistics for the experiments on the MNIST dataset}
    \label{fig:experiments}
\end{figure}

\begin{table}[!h]
\centering
\renewcommand{\arraystretch}{1.3}
\begin{tabular}{c|cc|c}
\toprule
Method & \textbf{RMSE} $\downarrow$ & \textbf{MAE} $\downarrow$ & \textbf{Cluster Acc. (in \%) $\uparrow$} \\
\midrule
DFedAvg & $0.165 \pm 0.008$ & $0.136$ & \textbf{---} \\
Ours ($p=2$)    & $\mathbf{0.140 \pm 0.008}$ & $\mathbf{0.109}$ & $85.3 \pm 5.3$ \\
\midrule
\textbf{$\Delta$} & \textbf{--15.3\%} & \textbf{--19.7\%} & --- \\
\bottomrule
\end{tabular}
\caption{Test performance on Milan+Trentino traffic forecasting, mean over
$K{=}20$ orbital snapshots. Our method improves error by $\sim$15--20\% over DFedAvg.}
\label{tab:one}
\end{table}

\begin{table}[!h]
\centering
\renewcommand{\arraystretch}{1.3}
\begin{tabular}{c|cc|c}
\toprule
Method & \textbf{Test Accuracy (in \%) $\uparrow$} & \textbf{Cluster Acc. (in \%) $\uparrow$} \\
\midrule
DFedAvg & $71.3 \pm 0.41$ & \textbf{---} \\
Ours ($p=2$)    & $\mathbf{80.8 \pm 0.13}$ & $100.0$ \\
\midrule
\textbf{$\Delta$} & $\mathbf{+9.5}$  & --- \\
\bottomrule
\end{tabular}
\caption{Test performance on MNIST, mean over
$K{=}20$ orbital snapshots. Our method improves test accuracy by $\sim$9.5\% over DFedAvg.}
\label{tab:two}
\end{table}

\subsection{Evaluation Metrics}

\subsubsection{Forecasting performance}
Forecasting performance is measured by RMSE and Mean Absolute Error (MAE) on a held-out test partition of cell time series. To account for the fact that each satellite's test-time cell assignment depends on the orbital snapshot, we report metrics averaged over $K = 20$ post-training snapshots, each drawn from an independent orbital configuration.

\begin{figure}[!h]
    \centering
    \includegraphics[width=\linewidth]{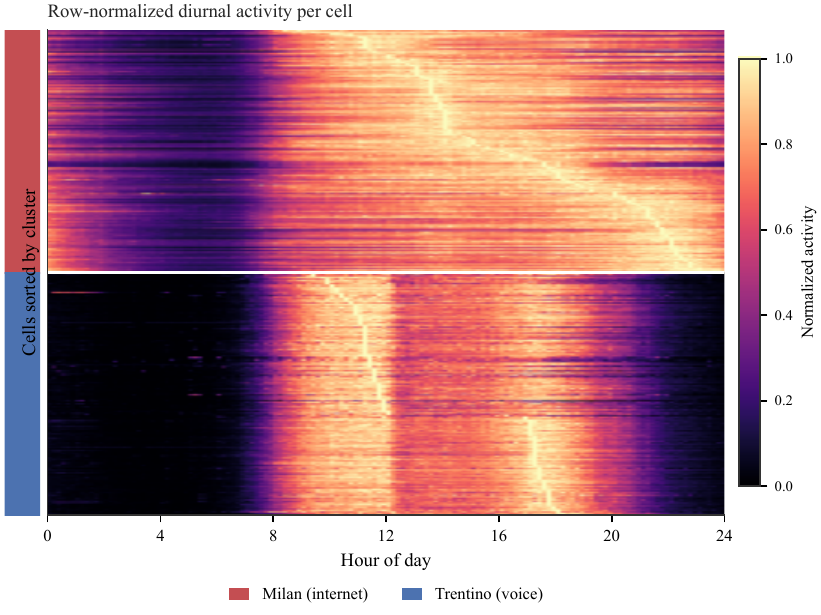}
    \caption{Heatmap visualizing heterogeneity of Milano+Trentino Telecommunications Dataset}
    \label{fig:data}
\end{figure}

\subsubsection{Clustering accuracy}
For the rotated MNIST experiments, we measure the cluster accuracy explicitly, judging based on the head selection. As the clustering for the Milan+Trentino Telecommunications dataset is implicit, as shown in Figure~\ref{fig:data}, we report cluster-recovery accuracy: the fraction of satellites whose selected head agrees with the latent regime label of the cell they observe.

\subsection{Results}
\label{sec:results}
Table~\ref{tab:one} reports test performance on the Milan+Trentino forecasting task, averaged over $K{=}20$ post-training orbital snapshots. The clustered extension reduces RMSE from $0.165$ to $0.140$ (a relative reduction of $15.3\%$) and MAE from $0.136$ to $0.109$ ($-19.7\%$) compared to a decentralized FedAvg baseline of identical per-satellite capacity. The gain is consistent across snapshots. The standard deviation of RMSE across the $K{=}20$ configurations is $0.008$ for both methods, so the improvement is not attributable to snapshot variance, a point we return to in Section~\ref{sec:discussion} when discussing the operational significance of this error reduction.  Cluster-recovery accuracy reaches $85.3\% \pm 5.3\%$, indicating that satellites converge to selecting the head aligned with the latent regime of the ground cell in their current footprint in a large majority of cases. Because the regime labels are constructed from the urban/rural partition of the ground cells (Section~\ref{sec:sys}), this accuracy is a proxy for how well the two learned heads correspond to the structural heterogeneity visible in Figure~\ref{fig:data} rather than to arbitrary partitions of the input space. The residual $\sim$15\% mismatch is consistent with boundary cells whose activity profile lies between the two regimes and for which either head is a reasonable local fit.  The MNIST rotation benchmark (Table \ref{tab:two} and Figure \ref{fig:experiments}) provides a controlled sanity check with a known ground-truth clustering. Under the same decentralized topology and $K{=}20$ snapshot protocol, our method reaches $80.8\% \pm 0.13$ test accuracy versus $71.3\% \pm 0.41$ for DFedAvg (a $9.5$ percentage-point improvement), while recovering the underlying rotation clusters with essentially perfect accuracy ($100\% \pm 0.05$). The fact that cluster recovery is near-perfect on MNIST but $\sim$85\% on Milan+Trentino reflects the difference between synthetic and natural heterogeneity. Rotation groups are cleanly separable in output space, whereas the urban/rural boundary in cellular traffic is statistical rather than categorical.  Taken together, the two experiments show that the clustered extension delivers measurable gains on both a controlled diagnostic (MNIST) and a realistic forecasting task (Milan+Trentino), and that the head-selection mechanism recovers meaningful structure in both cases.

\subsection{Discussion}
\label{sec:discussion}

\paragraph{Impact on network traffic prediction.} The headline result of this work is not merely that clustered specialization outperforms a shared model, but the scale of that improvement for a task where forecast error directly gates operational decisions. A $15.3\%$ RMSE reduction and $19.7\%$ MAE reduction are large relative to typical gains reported for architectural changes in cellular traffic forecasting, where single-digit percentage improvements are the norm. In a proactive resource-allocation pipeline, this error reduction does not translate linearly into cost or reliability. Forecasting errors sit on the tail of a provisioning decision. A satellite that under-predicts traffic must either drop or delay demand it did not provision for, while one that over-predicts wastes scarce onboard power and bandwidth reserving capacity that goes unused. Both failure modes are more expensive in the orbital setting than on terrestrial infrastructure, where a base station can draw on backhaul elasticity or neighboring cell handover that a satellite serving an isolated footprint does not have. A tighter forecast, concretely, means the provisioning margin a satellite must reserve against uncertainty can shrink without increasing the risk of a missed demand spike, freeing capacity that would otherwise sit idle as a safety buffer on every orbit. The consistency of the gain across $K=20$ independent orbital snapshots (Section~\ref{sec:results}) is what makes this actionable rather than incidental: an operator cannot tune a provisioning margin around an improvement that only appears under one topology, but a $0.008$ standard deviation across configurations means the $15$--$20\%$ error reduction is a property of the method, not of a favorable test split.

\paragraph{Further comments regarding the results of this research.}

\begin{enumerate}
    \item \textbf{Where the gain comes from:} Because DFedAvg and our method share the same per-satellite capacity, the observed RMSE reduction cannot be attributed to additional parameters at inference time, only one head is active per forward pass. The gain instead reflects that a single shared model is forced to compromise between the two regime shapes visible in Figure~\ref{fig:data}, whereas two specialized heads each fit one regime more tightly. The larger relative gain on MAE ($-19.7\%$) than on RMSE ($-15.3\%$) is consistent with this interpretation: MAE is more sensitive to the systematic bias a compromise model incurs on the tails of the two distributions, while RMSE partially averages that bias against high-variance peaks.  
    
    \item \textbf{Cluster recovery under natural heterogeneity:} The $\sim$15\% cluster-accuracy gap between MNIST and Milan+Trentino is expected. The MNIST rotation task is designed such that the two clusters are essentially disjoint in the loss landscape, so zero-shot head selection is unambiguous. Cellular traffic, in contrast, exhibits a continuum: Cells at the urban/rural boundary have profiles that neither head strictly dominates, and either assignment is locally near-optimal. That the algorithm nevertheless converges to the correct head in the large majority of cases, without any supervision on the regime labels, suggests that the structural heterogeneity is genuinely two-modal at the level the model can exploit.  
\end{enumerate}

\section{Conclusion}
\label{sec:conclusion}

\subsection{Conclusion}
\label{sec:conclusion-final}

We extended DFCA to the LEO satellite setting, where continuous orbital motion induces a time-varying local data distribution at each satellite and where ground-cell traffic patterns are structurally heterogeneous. Adapting the algorithm to a periodically rebuilt $k$-nearest inter-satellite gossip topology and evaluating under a $K{=}20$ snapshot protocol, we observed a $15.3\%$ RMSE reduction and $19.7\%$ MAE reduction on the Milan+Trentino cellular internet-traffic forecasting task, with cluster-recovery accuracy of $85.3\%$, and a $9.5$-percentage-point accuracy improvement with essentially perfect cluster recovery on the rotated MNIST control experiment.  

These results support the claim that clustered specialization is a useful mechanism in the orbital edge-compute setting, and that it can be realized without a ground-side aggregator through per-head zero-shot selection and same-head gossip aggregation. Extensions to more than two regimes, adaptive selection of the head count $p$, and joint modeling of communication cost alongside statistical performance are natural next steps for closing the gap between the diagnostic setup evaluated here and a deployable 6G non-terrestrial forecasting stack.

\subsection{Open Challenges} 
\label{sec:open-challenges}

First, our experiments use $p = 2$ regimes, matching the urban/rural structure of the Milan+Trentino dataset. Extending the evaluation to $p > 2$, to misspecified $p$, or to settings with more than two latent regimes is a promising direction, particularly to characterize how head selection degrades when the true number of regimes exceeds the model's capacity. Second, our evaluation treats the ground cell currently under each satellite as its data source. Incorporating beam footprints that span multiple cells, as in realistic LEO deployments, would be a valuable next step and may reveal how the per-satellite regime signal behaves under such blurring. Third, this work reports the statistical performance of the algorithm. A systems-level treatment that explicitly models communication cost, energy consumption, and ISL bandwidth constraints would complement these results and is left for future work.

\bibliography{references}
\bibliographystyle{plain}



\end{document}